\documentclass[letterpaper, 10 pt, conference]{ieeeconf} 

\IEEEoverridecommandlockouts 
\usepackage{graphicx}
\usepackage{amsmath}
\usepackage{arydshln}
\usepackage{hhline}
\usepackage{eucal}
\usepackage{cite}
\usepackage{amssymb}
\usepackage{enumerate}
\usepackage{subfigure}
\usepackage{bm}
\usepackage{color}
\usepackage{multirow}
\usepackage{graphicx}
\usepackage{svg}
\usepackage{booktabs}

\usepackage{cite}
\usepackage[linesnumbered,ruled,vlined]{algorithm2e}
\usepackage{amsmath}
\usepackage{amssymb}
\usepackage{xcolor}
\usepackage{comment}

\SetCommentSty{mycommfont}
\usepackage{graphicx}
\usepackage[symbol]{footmisc}
\usepackage{hyperref}

\DeclareSymbolFontAlphabet{\mathcal}   {symbols}

\title{\LARGE \bf
A Reachability Tree-Based Algorithm 
for Robot Task and Motion Planning
}

\author{Kanghyun Kim$^{1}$, Daehyung Park$^{2}$, and Min Jun Kim$^{1}$
\thanks{This work was supported by the National Research Foundation of Korea (NRF) grant funded by the Korea government (MSIT) No. 2021R1C1C1005232, and No. 2021R1A4A3032834}
\thanks{$^{1}$K. Kim, and M. J. Kim are with the School of Electrical Engineering, and $^{2}$D. Park is with the School of Computing, Korea Advanced Institute of Science and Technology (KAIST), Daejeon, Republic of Korea. E-mail: {\tt\small \{kh11kim, daehyung, minjun.kim\} at kaist.ac.kr}}
}

\begin{document}

\setlength{\textfloatsep}{0pt}

\maketitle
\thispagestyle{empty}
\pagestyle{empty}

\begin{abstract}
This paper presents a novel algorithm for robot task and motion planning (TAMP) problems by utilizing a reachability tree. While tree-based algorithms are known for their speed and simplicity in motion planning (MP), they are not well-suited for TAMP problems that involve both abstracted and geometrical state variables. To address this challenge, we propose a hierarchical sampling strategy, which first generates an abstracted task plan using Monte Carlo tree search (MCTS) and then fills in the details with a geometrically feasible motion trajectory. Moreover, we show that the performance of the proposed method can be significantly enhanced by selecting an appropriate reward for MCTS and by using a pre-generated goal state that is guaranteed to be geometrically feasible. A comparative study using TAMP benchmark problems demonstrates the effectiveness of the proposed approach. 
\end{abstract} 

\section{Introduction} \label{sec:introduction}

A number of studies are being conducted to enable robots to achieve autonomy and perform everyday tasks in an unstructured environment\cite{garrett2021integrated}. 
To this end, on top of low-level control algorithms, e.g., \cite{kim2021passive, jeong2022memory}, the robot should be able to plan an abstracted action sequence (by task planning/planner, TP), while ensuring the geometric feasibility of each action (by motion planning/planner, MP). It is worthwhile to note that the TP involves finding a discrete task-level action sequence, and the MP involves finding a collision-free path for a robot. Both TP and MP methods have already been studied extensively in the literature\cite{lavalle2006planning, ghallab2004automated}.

\begin{figure} [t!]
    \centering
	{\includegraphics[width=\linewidth]{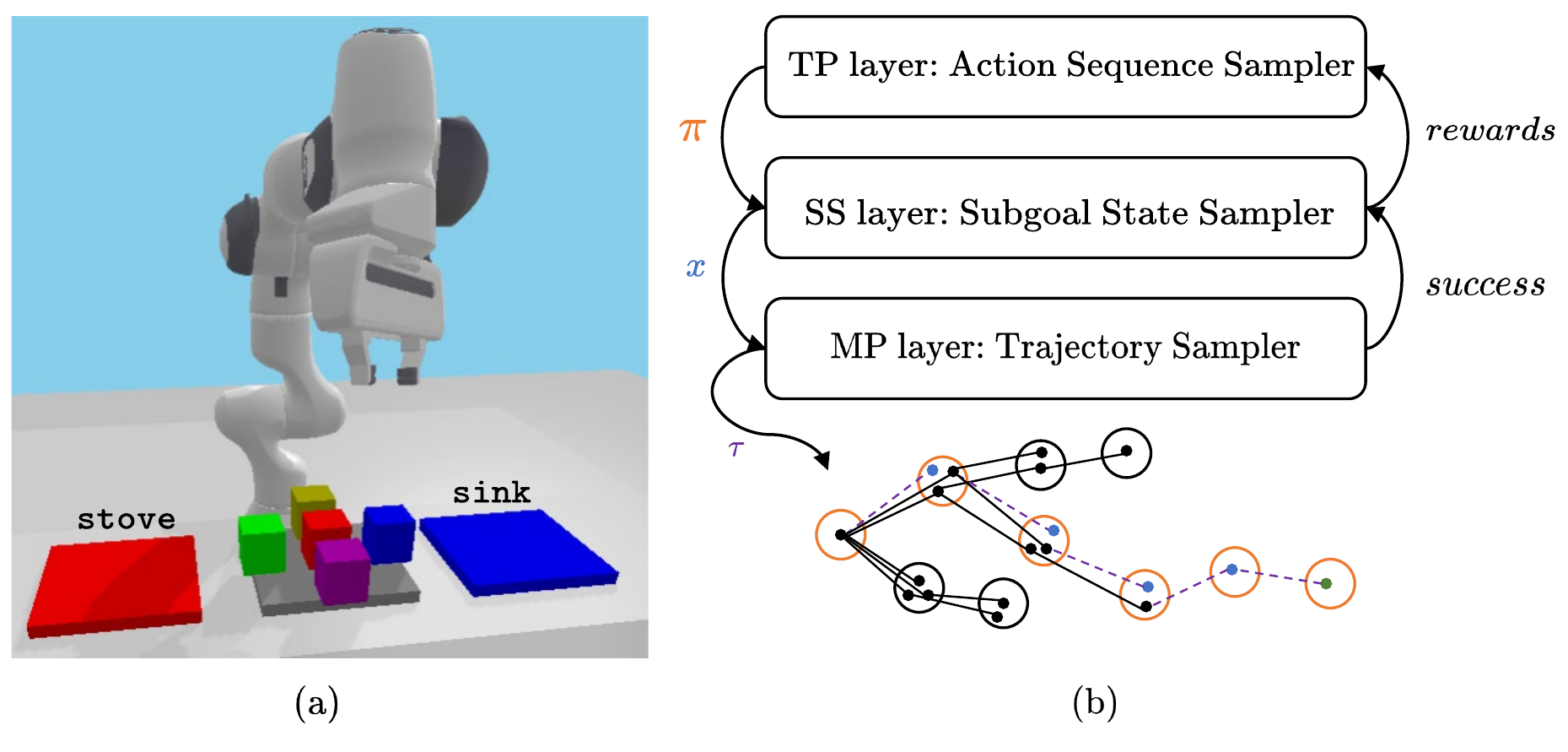}}
	    \vspace{-8mm}
    	\caption{Manipulation planning problems often involve tight geometric constraints as well as task-level planning. (a) In the kitchen domain, the robot needs to cook a given food block after washing it. (b) We propose an algorithm to build a reachability tree with hybrid states using three layers of planning hierarchies towards task and motion planning problems.}
	\label{fig:face}
    \vspace{1mm}
\end{figure}

To accomplish the robot manipulation task autonomously, both TP and MP problems need to be solved simultaneously, and this is referred to as task and motion planning (TAMP). Fig. \ref{fig:face} shows a motivating example of a TAMP problem called \textit{kitchen} domain. In this problem, the robot can have both geometric (e.g. joint angles, object poses) and abstracted (e.g. \texttt{cooked[object]}, \texttt{washed[object]}) state variables, and the TAMP planner should generate actions that satisfy not only geometric constraints (e.g. collision-free) but also task-level constraints defined in the problem (e.g. all food blocks must be washed before cooking). 
A major challenge of this setup is that, when the \texttt{sink} and/or \texttt{stove} have tight geometric constraints, the planner is likely to require a large amount of samplings.  TAMP problems generally become more complex when the geometry of the problem narrows the solution space. 
In this case, the high-level TP tends to suggest an action sequence that is not geometrically valid. Consequently, the low-level MP performs unnecessary collision checks, making the entire planner inefficient.

Our planning approach utilizes a \textit{reachability tree} that captures the connectivity of sampled states, like a rapidly-exploring random tree (RRT\cite{lavalle1998rapidly}). However, unlike MP problems, TAMP problems have a hybrid state space consisting of both abstracted and geometrical state variables, which poses a challenge in defining an effective distance metric between two states. For this reason, it is not trivial to utilize an RRT-like extension scheme, i.e. sampling a random node and extending the tree toward it, to TAMP problems.

To address this issue, we propose a hierarchical strategy that first samples an abstract action sequence by employing Monte Carlo tree search (MCTS) in conjunction with a symbolic planner (which can be of any off-the-shelf planner), and subsequently samples the corresponding geometric state variables. Moreover, we introduce two methods to improve planning speed. First, the reward of MCTS is generated by evaluating the result of TP (i.e. action sequence) using MP. Second, when performing MP, a goal state candidate is generated in advance based on the given action sequence. We found that prechecking the feasibility of the goal candidate can significantly improve the search speed, particularly in problems with tight goal constraints, such as the \textit{kitchen} domain. To validate the effectiveness of our approach, we compared it with our baselines and PDDLStream\cite{garrett2020pddlstream}.

\section{Related Work}
Unlike the typical MP problems, the overall system configuration of manipulation planning is extended due to the existence of unactuated objects. The topology of this problem has already been well-established by introducing a concept of a \textit{mode} that parametrizes varying collision-free space of the robot. This type of problem is called a \textit{multi-modal motion planning} (MMMP) \cite{hauser2010multi}. In addition to MMMP, the TAMP problem requires a more general setting that considers abstracted states and actions \cite{garrett2021integrated}.
Namely, despite the fact that MMMP and TAMP are both tackling manipulation planning problems, they have slightly different problem setups. Consequently, solution approaches to MMMP and TAMP have tended to be developed separately.

In the field of TAMP, a symbolic planner-based method is often used \cite{garrett2020pddlstream, garrett2018sampling, srivastava2014combined, lozano2014constraint, kaelbling2011hierarchical, dantam2018incremental}. In most cases, the continuous domain is pre-discretized or sampled online, and a symbolic planner using domain-independent heuristics is utilized to search discretized state space.
These methods are very general as they can plan even with abstract states and actions. However, if the continuous domain is not discretized densely, a valid trajectory of the plan may not be found. Conversely, if discretization is too dense, the size of the symbolic planning problem becomes very large. These issues may become problematic when geometric constraints are tight. Consequently, the overall planning easily becomes inefficient as the iteration continues.

In the field of MMMP, researchers often utilize a planner that directly handles geometry by retaining a reachability tree that represents the connectivity of sampled states \cite{mirabel2017manipulation, simeon2004manipulation, hauser2010multi, hauser2008motion, barry2013hierarchical}. One of the most well-known approaches in this area is the RRT-like MMMP method\cite{hauser2011randomized}, which samples a random state and extends the search tree towards that sample. Previous studies have shown that well-designed heuristics can improve search speed. \cite{hauser2011randomized} biases promising transitions using a utility table, which is calculated offline. In \cite{kingston2022scaling, kingston2020informing}, the edge weight of the \textit{mode transition graph} is updated using motion planning, and the result of the graph search is utilized to guide the tree extension. 

Inspired by the previous works, we propose a reachability-tree-based approach for TAMP problems in this paper.

\section{Manipulation Planning Problem}

We formulate a TAMP problem for robotic manipulation by combining MP, MMMP, and TP methods. In the following, we briefly define our problem first, and then explain the notations in detail.

\subsection{Problem Definition}
We define a manipulation planning problem as a tuple, $(x_I, \mathcal{X}_G)$, where $x_I\in\mathcal{X}$ is an initial state, and $\mathcal{X}_G$ is a set of goal states. The objective of this problem is to find a sequence of states $[x_0, x_1, ..., x_k]$, where $x_0=x_I$, $x_k \in \mathcal{X}_G$, and $k$ is the number of state transitions. We assume a quasi-static and fully observable system in this work.

It should be noted that a goal set $\mathcal{X}_G$ can be implicitly expressed as a set of values of some state variables, which is denoted as $G$. A goal set $\mathcal{X}_G$ is derived as a set of states, which can be made by conjunction of values in $G$, where unmentioned state variables may have arbitrary values.
\vspace{-1mm}

\subsection{Notations}

\subsubsection{Object $o$, Movable object $m$, Robot $r$} 
In TP, $\mathcal{O}$ is a set of objects that includes a set of movable objects $\mathcal{M}\subset \mathcal{O}$, and the robot $r\in\mathcal{O}$. Non-movable objects are also used for placing movable objects.

\subsubsection{Attachment $\alpha$} 
The kinematic relationship between the two objects is defined using an \textit{attachment}. An attachment $\alpha$ of a movable object $m$ can be defined as a tuple consisting of itself, the parent object, and the transformation between the two,
$$
\begin{gathered}
\alpha_m=(m, p, \, ^{m}T_{p}),       \\
\text{where} \quad m\in\mathcal{M}, \, p\in \mathcal{O},\, ^{m}T_{p}\in SE(3).
\end{gathered}
$$
If the kinematic parent $p$ is a robot $r$, the attachment is called \textit{grasp}; otherwise, it is called \textit{placement}.

\subsubsection{Mode $\sigma$} \label{mode}
In most manipulation problems, each movable object has one attachment (i.e., a kinematic tree). In this setting, a tuple of attachments of all movable objects in a certain state is called a \textit{mode}:
$$\sigma=\{\alpha_m|m\in \mathcal{M}\}\in\Sigma,$$
where the mode space $\Sigma$ is an infinite set of modes. Once the mode is determined, all unactuated degrees of freedom in the system are fixed. Namely, a mode defines the collision-free configuration space of a robot, which we denote as $\mathcal{C}^{\sigma}_{free}$.

\subsubsection{Abstract state $s$} 
The planning domain definition language (PDDL)\cite{mcdermott1998pddl} can be used to abstract a state of the system, which we denote as $s$. Among variables that compose an abstract state, there are abstract attachments in the following form,
$$\tilde{\alpha}_m=\texttt{(attached ?m ?p)},$$
that indicates which object is attached to which parent. Note that the abstract version of an attachment simply omits transformation information $mT_p$ between two objects. Likewise, we can also naturally define an abstract mode $\tilde{\sigma}$. Conversely, the set of variables that is not abstract attachment is defined as non-geometric state $\psi$. Therefore, $s=(\psi, \tilde{\sigma}).$

\subsubsection{Abstract action $a$} 
The abstract action space $\mathcal{A}$ is a set of actions $a$, which can make a transition from an abstract state $s$ when applicable. We denote an applicable transition as, $$s'=T(s, a).$$ Similar to the abstract state, we call the special type of action $a_g \in \mathcal{A}$, which can change abstract mode $\tilde{\sigma}$ a geometric action, i.e. \texttt{Pick} and \texttt{Place}. Non-geometric action $a_n \in \mathcal{A}$ can only change non-geometric variables, such as \texttt{wash} or \texttt{cook} in \textit{kitchen} domain (see \ref{sec:eval}). A set of geometric actions and a set of non-geometric actions are denoted by $\mathcal{A}_g$ and $\mathcal{A}_n$, respectively.

\begin{figure} [tb!]
    \centering
	{\includegraphics[width=0.9\linewidth]{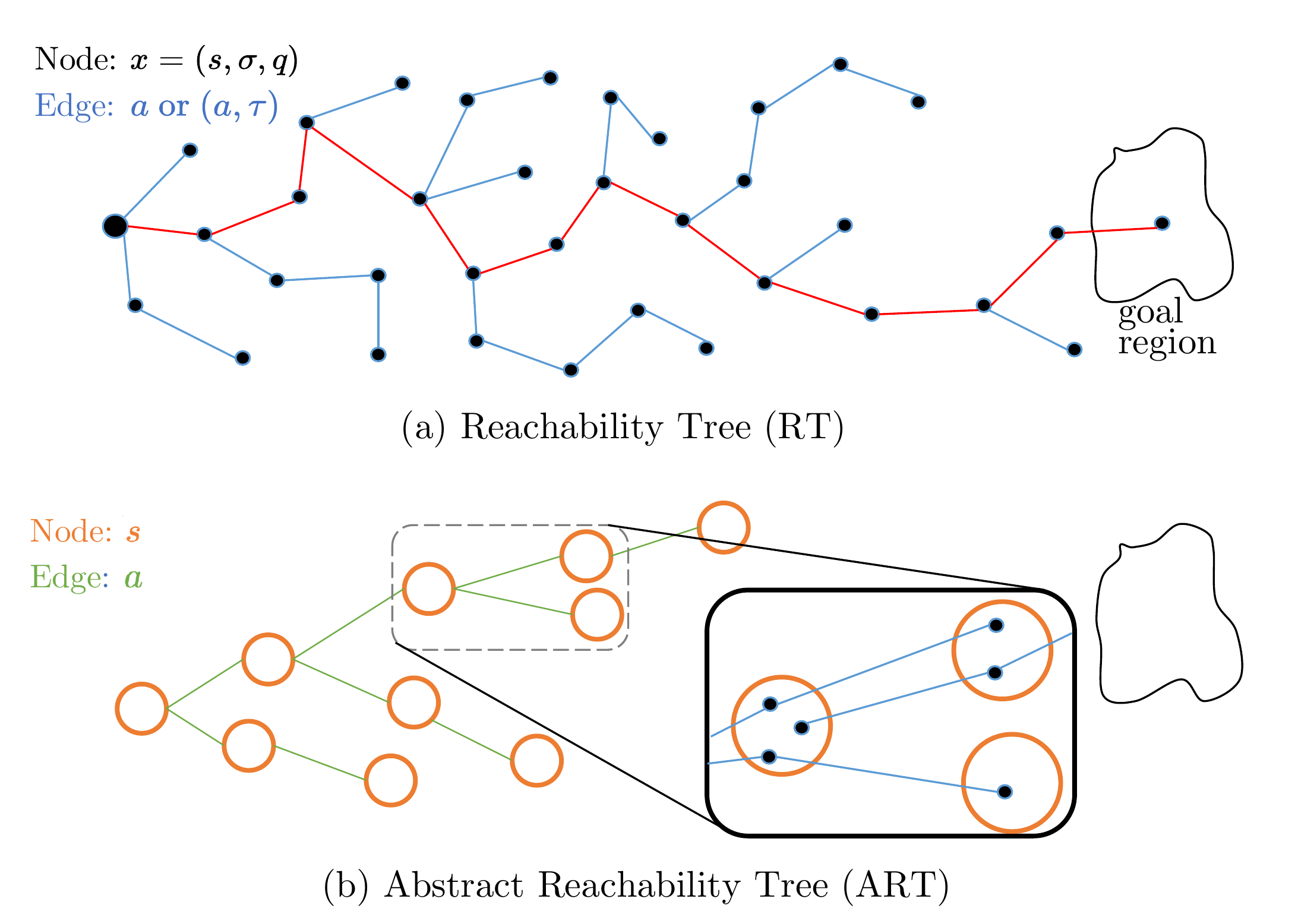}}
	    \vspace{-3mm}
    	\caption{Our algorithm maintains two trees: (a) The reachability tree (RT) tracks all reachable states from the initial state, and (b) the abstract reachability tree (ART) is an abstraction of the reachability tree, with each node containing multiple RT nodes.}
	\label{fig:tree}
 	\vspace{1mm}
\end{figure}

\subsubsection{Hybrid state $x$}\label{subsubsec:state} 
Now we can represent the full state of the system, $$x\in \mathcal{X} = \Psi \times \Sigma \times \mathcal{C} \, ,$$ where $q$ is a configuration of the robot, and $\Psi, \Sigma, \mathcal{C},$ and $\mathcal{X}$ are non-geometric state space, mode space, robot configuration space, and hybrid state space, respectively. 
For the sake of simplicity, we will use the redundant representation as,
$$x=(s, \sigma, q),$$
where the information of the abstract mode is redundantly expressed in both $s$ and $\sigma$.

\subsubsection{State transition}\label{statetransition} There are 3 types of state transitions between two states $x=(s, \sigma, q)$ and $x'=(s', \sigma', q')$, and each transition induces a three-level hierarchical structure in our planner.
\begin{itemize}
    \item \textit{Configuration transition}: Only $q$ is changed. That is, $$||q'-q||<\epsilon, \enspace \sigma=\sigma', s=s',$$ where $\epsilon\in\mathbb{R}$ is a small number which implies motion continuity.
    \item \textit{Mode transition}: $\sigma$ can be changed only if $q$ is in the subspace induced by two modes, $$\sigma\neq \sigma', \enspace q=q'\in\mathcal{C}_{free}^{\sigma} \cap \mathcal{C}_{free}^{\sigma'},  \enspace \psi_s=\psi_{s'},$$ where $s'=T(s, a_g), \enspace a_g\in\mathcal{A}_g$. $\phi_s$ and $\phi_{s'}$ represents the nongeometric states of $s$ and $s'$, respectively.
    \item \textit{Non-geometric state transition}: Only non-geometric state $\psi$ is changed, therefore, $$\psi_s \neq \psi_{s'} \enspace q=q', \enspace \sigma = \sigma',$$ where $s'=T(s, a_n), \enspace a_n\in\mathcal{A}_n.$
\end{itemize}

\section{Our Planning Algorithm$^*$}
\footnotetext[1]{The corresponding video and code are available on the project website: \url{https://sites.google.com/view/tree-based-tamp}}
\footnotetext[1]{Note that the . (dot) operator is used to indicate a member of a node for algorithm description.}

\begin{figure} [tb!]
    \centering
	{\includegraphics[width=0.95\linewidth]{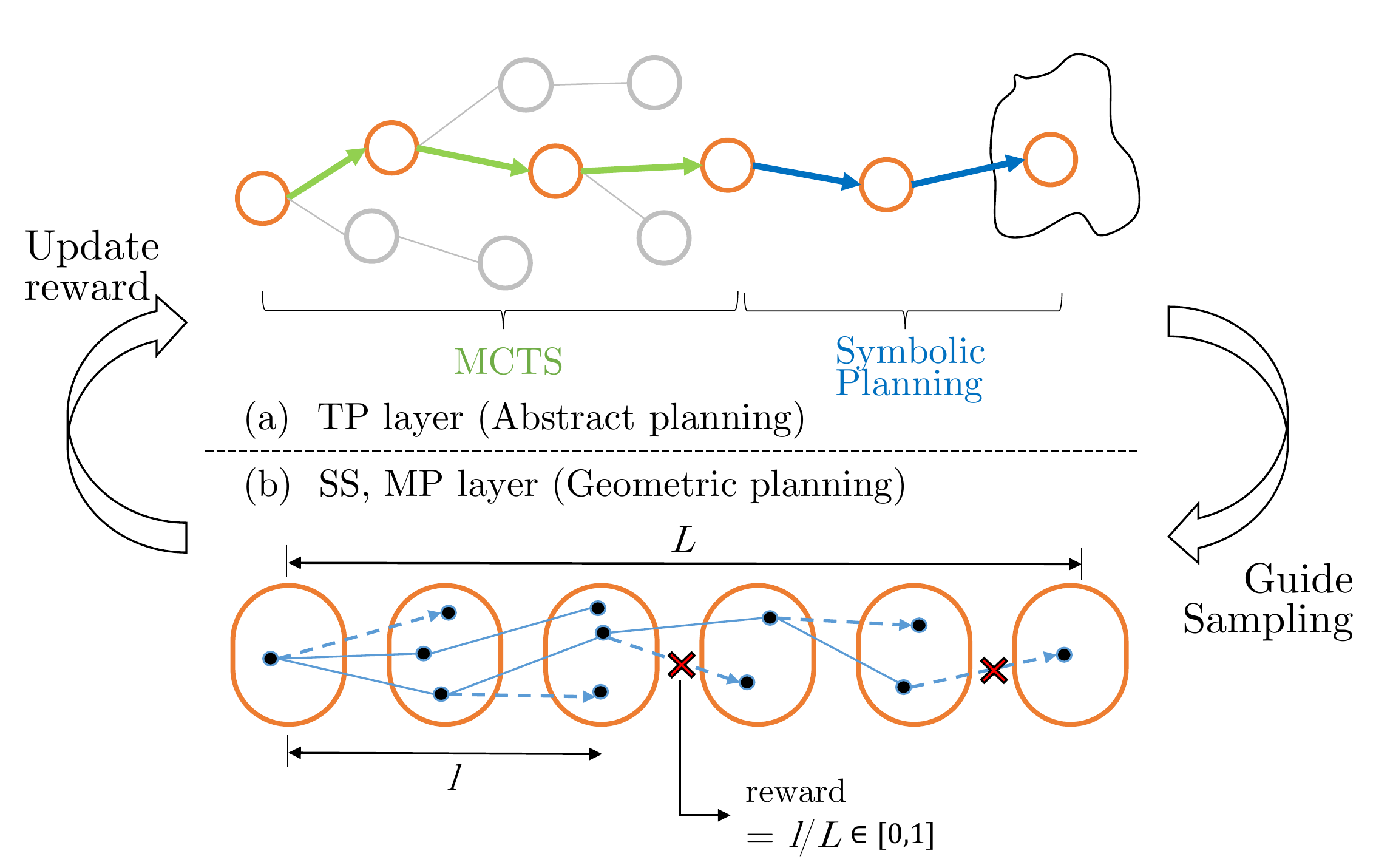}}
	    \vspace{-3mm}
    	\caption{(a) TP layer utilizes MCTS and a symbolic planner to sample an action sequence. (b) The RT extension is guided by the sampled action sequence. When the extension fails, the SS layer calculates a reward.}
	\label{fig:method}
 	\vspace{1mm}
\end{figure}

Our algorithm expands a reachability tree (RT) until the goal state is included in the tree (Fig.~\ref{fig:tree}(a)). The key idea is to use sampling of action sequences (i.e. abstract plans) to expand the RT. To accomplish this, we use the \textit{abstract reachability tree} (ART), which is an abstraction of the RT (Fig.~\ref{fig:tree}(b)). Since ART only considers task-level information, one ART node naturally includes multiple RT nodes, as depicted by the black solid box in Fig. \ref{fig:tree}(b).

Fig. \ref{fig:method} illustrates an overview of the entire algorithm. The tree expansion is performed hierarchically using three layers: (1) action sequence sampling by the TP layer, (2) subgoal sampling (SS) by the SS layer, and (3) trajectory sampling by the MP layer. 

First, in the TP layer, an action sequence is sampled using only ART. The action sequence sampler consists of a randomized tree search and a symbolic planner. That is, a partial plan is obtained by the randomized tree search and the remaining plan is completed using a symbolic planner (Fig. \ref{fig:method}(a)). Note that using MCTS as a randomized tree search naturally leads to informed sampling in the TP layer. Second, in the SS layer, new RT nodes for exploration are sampled based on the abstract plan, which is the outcome of the TP layer. Note that attachment sampling is required for each abstract action (e.g. \texttt{pick} requires a grasp). The SS layer samples the attachments in a batch manner to create a goal state candidate in advance. If the candidate is infeasible, all batch attachments are rejected. Otherwise, the sampled attachments are used to create new RT nodes. Lastly, the MP layer samples the trajectory between a tree node and a new node. If the MP fails, the SS layer calculates the ratio of the successful extension length of an abstract plan as a reward (Fig. \ref{fig:method}(b)). The average of rewards is then updated to ART and utilized to bias the sampling of better abstract plans in the next iteration.

A formal description of the two main trees is as follows.
\begin{itemize}
  \item \textit{Reachability tree (RT)} $\mathcal{G}$: Each node corresponds to a hybrid state $x$, and each edge represents the transition between two states. For geometric actions, the edge also includes a trajectory, while for non-geometric actions, the edge only includes the action itself.
    
  \item \textit{Abstract reachability tree (ART)} $\mathcal{T}$: Each node corresponds to an abstract state $s$, and each edge represents an abstract action $a$. In addition, each node contains information such as the number of visits $n_{visit}$, the sum of rewards $r_{total}$, and a set of RT node pointers $V_s$. Note that $n_{visit}$ and $r_{total}$ are used in the MCTS in the TP layer, while $V_s$ is used for parent sampling in the SS layer.
\end{itemize}

\subsection{Task Planning Layer (Alg. \ref{tplayer})}

\begin{algorithm}[t]
\caption{Task Planning Layer}\label{tplayer}
\small
\DontPrintSemicolon
\SetKwProg{Fn}{Function}{:}{}
\SetKwFunction{RG}{BuildRG}
\SetKwFunction{ReachabilityGraph}{ReachabilityTree}
\SetKwFunction{HashTable}{HashTable}
\SetKwFunction{SampleActionSeq}{SampleActionSeq}
\SetKwFunction{GetGeomStateSeq}{GetGeomStateSeq}
\SetKwFunction{AbstractStateTree}{AbstractStateTree}
\SetKwFunction{SSLayer}{SSLayer}
\SetKwFunction{AddNode}{AddNode}
\SetKwFunction{IsSolutionFound}{IsSolutionFound}
\SetKwFunction{UpdateTree}{UpdateTree}
\SetKwFunction{Queue}{Queue}
\SetKwFunction{Concat}{Concat}
\SetKwFunction{Average}{Average}
\SetKwFunction{GetAbsStateNodeSeq}{GetAbsStateNodeSeq}
\SetKwInOut{Input}{Input}
\SetKwInOut{Output}{Output}

\Input{Initial state $x_I$, Implicitly defined goal set $G$, \quad
SS layer Iteration number $k_{ss}$}
\Output{Reachability Tree $\mathcal{G}$}
$\mathcal{G} \leftarrow$ \ReachabilityGraph{} \;
$\mathcal{T} \leftarrow$ \AbstractStateTree{} \;
$\mathcal{G}.$\AddNode{$x_I$} \;
$\mathcal{T}.$\AddNode{$x_I.s$} \;
\While{within computational budget}{
    $\boldsymbol{r} \leftarrow \Queue{}$\;
    $\boldsymbol{\pi} \leftarrow $ \SampleActionSeq{$\mathcal{T}, G$}\tcp*{Alg.\ref{acseqsampler}}
    $\boldsymbol{n_s} \leftarrow$ \GetAbsStateNodeSeq{$\mathcal{T}, \boldsymbol{\pi}$} \;
    \For{$i \leftarrow 1$ \KwTo $k_{ss}$}{
        $rewards \leftarrow $\SSLayer{$\boldsymbol{\pi}, \boldsymbol{n_s}, G,  \mathcal{G}$}\tcp*{Alg.\ref{algss}} 
        $\boldsymbol{r} \leftarrow $\Concat{$\boldsymbol{r}, rewards$}
    }
    \UpdateTree{$\boldsymbol{n_s}$, \Average{$\boldsymbol{r}$}}
}
\end{algorithm}

The task planning layer is shown in Alg. \ref{tplayer} and Fig. \ref{fig:method}(a). First, RT $\mathcal{G}$ and ART $\mathcal{T}$ are initialized (line 1-4). Within the loop, \SampleActionSeq (Alg. \ref{acseqsampler}) outputs a sequence of actions $\boldsymbol{\pi}$, ensuring that the initial state can be successfully transitioned to the abstract goal state set. Note that the set of abstract goal states can be derived by the implicitly defined goal set $G$ (line 7). Moreover, a node sequence of abstract states $\boldsymbol{n_s}$ can also be derived using $\boldsymbol{\pi}$ (line 8). This sampled sequence $\boldsymbol{\pi}$ and $\boldsymbol{n_s}$ are evaluated $k_{ss}$ times by the subgoal sampling layer (line 10). The ART $\mathcal{T}$ is updated using the average of rewards obtained by the SS layer.

\subsection{Action Sequence Sampler (Alg. \ref{acseqsampler})} \label{sub:ActionSeqSamp}

\begin{algorithm}[t]
\caption{Action Sequence Sampler}\label{acseqsampler}
\small
\DontPrintSemicolon
\SetKwProg{Fn}{Function}{:}{}
\SetKwFunction{ActionSeqSampler}{ActionSeqSampler}
\SetKwFunction{RandomizedTreeSearch}{RandomizedTreeSearch}
\SetKwFunction{Concat}{Concat}
\SetKwFunction{ExtendTree}{ExtendTree}
\SetKwFunction{TaskPlanner}{TaskPlanner}
\SetKwInOut{Input}{Input}
\SetKwInOut{Output}{Output}
\Input{Abstract reachability tree $\mathcal{T}$, Implicitly defined goal set $G$}
\Output{Action sequence $\boldsymbol{\pi}$}

$n_{last}, \boldsymbol{\pi_{rand}} \leftarrow $\RandomizedTreeSearch{$\mathcal{T}$} \;
$\boldsymbol{\pi_{plan}} \leftarrow$ \TaskPlanner{$n_{last}, G$} \;
\ExtendTree{$\mathcal{T}, n_{last}, \boldsymbol{\pi_{plan}}$}\;
$\boldsymbol{\pi} \leftarrow$ \Concat{$\boldsymbol{\pi_{rand}}, \boldsymbol{\pi_{plan}}$} \;
\end{algorithm}

The Action Sequence Sampler should provide a randomized sequence of abstract actions that can guide RT $\mathcal{G}$ to the goal set $\mathcal{X}_G$. The key idea is to create a partial action sequence using a randomized tree search, where the remaining plan can be completed using a symbolic planner. 
  
For \texttt{RandomizedTreeSearch}, we use a mechanism of Monte Carlo tree search (MCTS), which continues to evaluate the value of each node and outputs a better state sequence in the next iteration (line 1). In such a framework, it is important to balance between exploration for searching for better rewards and exploitation that makes use of current information. For this purpose, we employ a $\epsilon$-greedy strategy in node selection. That is, the child node is selected with probability $\epsilon$, otherwise, the child with the highest value is chosen. Because our lower-level planners are semi-complete, it should be possible to resample the same action sequence. For this reason, each tree node has a probability of terminating the sequence $\boldsymbol{\pi}$. This allows the \texttt{RandomizedTreeSearch} to resample the previously sampled sequence. As a result, the \texttt{RandomizedTreeSearch} outputs the randomized plan $\boldsymbol{\pi_{plan}}$ and the final ART node $n_{last}$.
  
\texttt{TaskPlanner} can be any off-the-shelf symbolic planner (line 2). Then, the planned sequence $\boldsymbol{\pi_{plan}}$ is used to expand the ART $\mathcal{T}$ (line 3). This increases the probability of selecting a similar sequence to the current sequence $\boldsymbol{\pi}$ in the next iterations.

\subsection{Subgoal Sampling Layer (Alg. \ref{algss})} \label{mmplayer}

\begin{algorithm}[t!]

\caption{Subgoal Sampling Layer}\label{algss}
\small
\DontPrintSemicolon
\SetKwProg{Fn}{Function}{:}{}

\SetKwFunction{SampleParents}{SampleParents}
\SetKwFunction{GetGeomStateSeq}{GetGeomStateSeq}
\SetKwFunction{GetTreeNodeSeq}{GetTreeNodeSeq}
\SetKwFunction{Queue}{Queue}

\SetKwFunction{AddEdge}{AddEdge}
\SetKwFunction{SampleGoalState}{SampleGoalState}
\SetKwFunction{SampleRGNode}{SampleRTNode}
\SetKwFunction{GetAbsStateNodeSeq}{GetAbsStateNodeSeq}
\SetKwFunction{GetGoalMakingActionSet}{GetGoalMakingActionSet}
\SetKwFunction{IsGoalMakingAction}{IsGoalMakingAction}
\SetKwFunction{SampleTransition}{SampleTransition}
\SetKwFunction{SampleSingleModePath}{SampleSingleModePath}
\SetKwFunction{RegisterRGNode}{RegisterRGNode}
\SetKwFunction{GetAdjacentModeToGoal}{GetAdjacentModeToGoal}
\SetKwFunction{SampleAdjacentMode}{SampleAdjacentMode}
\SetKwFunction{RegisterRGNodePtr}{RegisterRGNodePtr}
\SetKwFunction{SampleBatchAtt}{SampleBatchAttachments}
\SetKwFunction{MakeGoal}{MakeGoalCandidate}
\SetKwFunction{Push}{Push}
\SetKwFunction{MPLayer}{MPLayer}
\SetKwFunction{RG}{RG}

\SetKwFunction{Average}{Average}
\SetKwFunction{MakeNextMode}{MakeNextMode}

\SetKwInOut{Input}{Input}
\SetKwInOut{Output}{Output}
\Input{Action sequence $\boldsymbol{\pi}$, Abstract state node sequence $\boldsymbol{n_s}$, Implicitly defined goal set $G$, Reachability tree $\mathcal{G}$, Iteration number of goal candidate sampling $k_{goal}$}
\Output{A sequence of rewards $rewards$}

\tcp{Goal candidate generation}
$x_I \leftarrow \mathcal{G}.root$\;
\For{$i \leftarrow 1$ \KwTo $k_{goal}$}{
    $\boldsymbol{\alpha}\leftarrow $\SampleBatchAtt{$\boldsymbol{\pi}, G$}\tcp*{Alg.4}
    $x_G\leftarrow$ \MakeGoal{$x_I, \boldsymbol{\alpha}$}\;
    \lIf{there is no collision at $x_G$}{break}
    $x_G\leftarrow$\texttt{null}
}
\lIf{$x_G\leftarrow$\texttt{null}}{\KwRet}

\tcp{Batch Extension of RT}
$rewards \leftarrow$ \Queue{}\;
\For{$i\leftarrow 1$ \KwTo $\boldsymbol{\pi}.length$}
{
    $n \leftarrow \boldsymbol{n_s}[i];$\enspace $a\leftarrow \boldsymbol{\pi}[i];$\enspace $\alpha \leftarrow \boldsymbol{\alpha}[i]$; \enspace $n'\leftarrow\boldsymbol{n_s}[i+1]$\; 
    $x \leftarrow $\SampleRGNode{$n$} \tcp*{Select parent}
    \lIf{$x = $\texttt{null}}{break}
    $s \leftarrow n.s; \enspace\sigma \leftarrow x.\sigma;$\enspace $q \leftarrow x.q\enspace s' \leftarrow n'.s$
    \;\;
    \If{$a\notin\mathcal{A}_g$}
    {
        \tcp{Non-geometric transition}
        $x' \leftarrow (s', \sigma, q)$ 
    }
    \Else{
        \tcp{Mode transition}
        $\sigma'\leftarrow $\MakeNextMode{$\sigma, \alpha$}\;
        $q'\leftarrow$ \SampleTransition{$\sigma, \sigma'$}\;
        $x' \leftarrow (s', \sigma', q')$
    }
    $success \leftarrow $ \MPLayer{$x, x', \mathcal{G}$} \;
    \lIf{$success$}{$n'.V_s \leftarrow n'.V_s \cup \{x'\}$
    }
    \lElse(\tcp*[f]{Reward}){$rewards.$\Push{$(i-1)/\boldsymbol{\pi}.length$}}
}
\If{the last tree extension was successful}{
    $success \leftarrow $\MPLayer($x', x_G, \mathcal{G}$)\;
    \If{$success$}{
    $\mathcal{G}.solution \leftarrow x_G$\tcp*{Solution found}
    }
}
\end{algorithm}

In the SS layer, the expansion of the RT is achieved through three steps. Firstly, this layer performs batch sampling of attachments required to realize the sampled action sequence. Secondly, new transition states are generated using the tree nodes and the sampled attachments. Lastly, a reward signal is generated to provide information to the TP layer.
 
The SS layer first samples all attachments for tree extension in a batched manner, and generates a candidate goal state $x_G$ based on the batch samples (lines 1-7). The attachment samples are rejected if the goal state candidate is not in collision. Detailed explanations of \texttt{SampleBatchAtt} will be introduced in \ref{SecGoalSampler}.
\MakeGoal makes a goal state candidate $x_G=(s_G, \sigma_G, q_G)$, where
\begin{itemize}
    \item $s_G$ can be derived by applying action sequence $\boldsymbol{\pi}$ to the initial abstract state $x_I.s$;
    \item $\sigma_G$ can be generated by applying mode changes by a sequence of attachments $\boldsymbol{\alpha}$ to the initial mode $x_I.\sigma$;
    \item $q_G$ can be found in $G$ or randomly sampled.
\end{itemize}
If $x_G$ is collision-free, attachments and goal states are used for the further tree extension process. Since sampling a goal candidate is computationally cheap, unachievable attempts can be rejected by validating the geometric feasibility of $x_G$.

After sampling the goal candidate, a batch extension of RT is conducted using $\boldsymbol{\pi}, \boldsymbol{n_s}$, and $\boldsymbol{\alpha}$. Fig. \ref{fig:method}(b) shows the batch extension process of RT. First, a parent RT node $x$ for expansion is sampled from the abstract state node $n$ (line 12). If $a$ is a non-geometric action, the new state $x'$ is created by changing the abstract state $s$ to $s'$, which is a non-geometric state transition (lines 16-17). If $a$ is a geometric action, an adjacent mode $\sigma'$ is generated by changing an attachment $\alpha$ from the previous mode $\sigma$ (line 19), and the transition configuration $q$ between $\sigma$ and $\sigma'$ is sampled (line 20). Note that this robot configuration sampling can be achieved by projecting a random configuration into a subspace induced by the two modes $\sigma$ and $\sigma'$, i.e. numerical inverse kinematics.

The connection between the new node $x'$ and the parent node $x$ is checked through the MP layer (line 22). If connected, the pointer of $x'$ is stored in the next abstract state node $s'$ (line 23). If the connection fails, a reward signal is calculated (line 24). We define a reward as the total extension length from the root to the current node, which is scaled by the length of the action sequence to have a value between 0 and 1. Note that multiple rewards are received through the batch extension process, and therefore, the TP layer is updated using the average of received rewards.
  
If the last action of the action sequence $\boldsymbol{\pi}$ is successful, the SS layer finally checks a connection between the last extended state $x'$ and the goal candidate $x_G$ (lines 25-26). If the extension to the goal candidate $x_G$ succeeds, a solution is found (lines 17-18). The path from the initial state $x_I$ to the goal state $x_G$ can be derived by backtracking the RT $\mathcal{G}$.

\subsection{Attachment Batch Sampler (Alg. \ref{algabs})} \label{SecGoalSampler}

\begin{algorithm}[t]
\caption{Attachment Batch Sampler}\label{algabs}
\small
\DontPrintSemicolon

\SetKwInOut{Input}{Input}
\SetKwInOut{Output}{Output}

\SetKwFunction{GetRandomConfig}{GetRandomConfig}

\SetKwFunction{SampleAttachment}{SampleAttachment}
\SetKwFunction{Queue}{Queue}
\SetKwFunction{Push}{Push}
\SetKwFunction{Reverse}{Reverse}
\SetKwFunction{MakeNextMode}{MakeNextMode}
\SetKwFunction{TargetObject}{TargetMovable}
\SetKwFunction{NextParent}{NextParent}
\SetKwFunction{IsCollision}{IsCollision}
\SetKwFunction{GetGoalAttachment}{GetGoalAttachment}
\SetKwFunction{IsGoalAttachment}{IsGoalAttachment}
\SetKwFunction{GetAttachmentOfObject}{GetAttachmentOfObject}

\Input{Action sequence $\boldsymbol{\pi}$, Implicitly defined goal set $G$}
\Output{A sequence of attachments $\boldsymbol{\alpha}$}

$\boldsymbol{\alpha} \leftarrow$ \Queue{}\;
$\mathcal{M}_{goal} \leftarrow \mathcal{M}$\;
\For{$i \leftarrow \boldsymbol{\pi}.\textit{length}$ \KwTo $1$}{
    $a \leftarrow \boldsymbol{\pi}[i]$\;
    \If{$a \notin \mathcal{A}_g$}{
        $\boldsymbol{\alpha}.$\Push{\texttt{null}}\;
        continue
    }
    $m \leftarrow $\TargetObject{a}\;
    \If{$m\in \mathcal{M}_{goal} \textbf{ and }$ \IsGoalAttachment{$a, G$}}{ 
        $\alpha_m \leftarrow \GetGoalAttachment{a, G}$
    }
    \Else{
        $\alpha_m \leftarrow$ \SampleAttachment{$a$}\;
    }
    $\boldsymbol{\alpha}.$\Push{$\alpha_m$}\;
    $\mathcal{M}_{goal} \leftarrow \mathcal{M}_{goal} - \{m\}$
}
$\boldsymbol{\alpha} \leftarrow $\Reverse{$\boldsymbol{\alpha}$}\;

\end{algorithm}

The attachment batch sampler samples all necessary attachments in the action sequence $\boldsymbol{\pi}$ at once. It is important to note that to complete the planning, the specific actions must select the corresponding attachments in the implicitly defined goal set $G$. We denote this action and attachment as \textit{goal-making action} and \textit{goal attachment}. Determining the goal-making actions can be achieved by examining the action sequence in reverse order and identifying the action that leads to the final mode. To accomplish this, a set $\mathcal{M}_{goal}$ is employed to keep track of the movables whose final attachments are not determined.
  
This attachment assignment procedure ignores non-geometric actions, which do not require attachments (lines 5-7). If $a$ is a geometric action, then it is necessary to determine its corresponding attachment. First, we obtain the target object $m$ from action $a$. If there is no attachment assigned for the object $m$ and if the action $a$ has a corresponding goal attachment in the implicitly defined goal set $G$, then it can be concluded that the action $a$ is the goal-making action. In this case, instead of sampling, we simply choose the goal attachment from the set $G$ (lines 9-10). If the action is not directly related to the goal, the attachment is sampled based on the action (lines 11-12). 

\section{Empirical Evaluation}\label{sec:eval}

\begin{figure} [t!]
    \centering
	{\includegraphics[width=\linewidth]{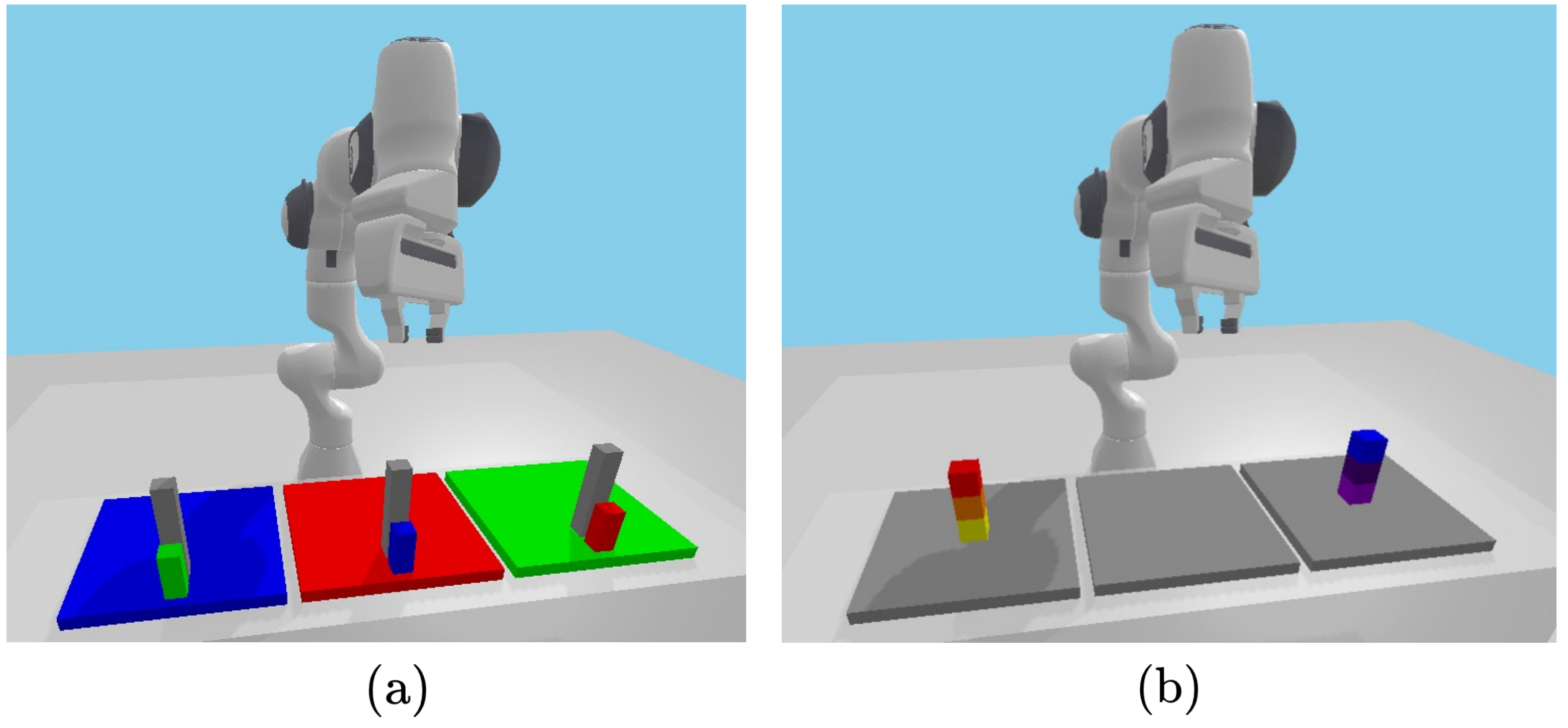}}
	    \vspace{-7mm}
    	\caption{(a) Non-monotonic domain. (b) Blocktower domain.}
	\label{fig:domains}
 	\vspace{1mm}
\end{figure}

This section validates the proposed method using three TAMP benchmark problems:
\begin{itemize}
\item{\textit{Kitchen $m$ domain\cite{ren2021extended}}:}
We provide $m$ number of movable food blocks and three placeable objects (i.e., \texttt{dish}, \texttt{sink}, and \texttt{stove}) as shown in Fig. \ref{fig:face}. The task is to cook every food block, which is initially placed on the \texttt{dish}. A food block must be \texttt{cleaned} before being \texttt{cooked}. Therefore, there are two non-geometric actions \texttt{cook} and \texttt{wash} that can be executed when a food block is placed on the \texttt{stove} and \texttt{sink}, respectively. This domain has a tight geometric constraint, i.e., the area of the \texttt{sink} and \texttt{stove} is small, so the blocks can only be placed densely.
\item{\textit{Non-monotonic $m$ domain\cite{lagriffoul2018platform}}:}
We provide $m$ blocks of different colors and $m$ blockers with a higher height. The robot task is to place colored blocks on the floor which must have the same color as the blocks. As shown in Figure \ref{fig:domains}(a), a difficulty arises since the robot cannot reach the color blocks at once. Therefore, grey blockers must be removed before approaching the color blocks. However, since the task planner is not aware of such geometry, so interaction between the TP layer and the lower-level layers is essential.
\item{\textit{The blocktower $m$ domain\cite{khodeir2021learning}}:}
We provide $m$ blocks and 3 placeable plates. Initially, we stacked blocks in random order on the left or right plates, as shown in Fig. \ref{fig:domains}(b). The robot task is to stack them on the center plate in a certain order. To account for the task constraint where a movable object can be placed on another movable object, two abstract actions, namely, \texttt{stack} and \texttt{unstack}, are introduced.
\end{itemize}

For comparison, we implemented PDDLStream, which is the most well-known and readily available TAMP solver online. In addition, two baseline planners are designed:
\begin{itemize}
\item{No-reward planner:} Our planner uses the reward as a bias for better action sequence sampling. To verify the effectiveness of this, the reward is not used in this baseline.
\item{No-rejection planner:} Our planner rejects sample batches by making a goal candidate and checking its geometrical feasibility. To verify the effectiveness of this process, the feasibility of the goal state is not checked in this baseline.
\end{itemize}

All planners including PDDLStream were written in Python. For the proposed method and baseline planners, the abstract domain is modeled in PDDL, and the task plan is solved by Pyperplan symbolic planner \cite{alkhazraji-et-al-zenodo2020}. For the MP layer, our implementation of RRTConnect\cite{kuffner2000rrt} is used. Collision checking and inverse kinematics are implemented using Pybullet physics library\cite{coumans2016pybullet}. All algorithms were executed on an Intel Core i7-10700 CPU, 16Gb RAM PC. 
The following hyperparameters were used in the proposed method including the baselines: $k_{ss}$ = 2, $k_{goal} = 10$, and $\epsilon = 0.5$. For PDDLStream, \textit{adaptive}, which is known as the best-performing implementation, was used with default parameters. All experiments were performed 30 times, and the timeout was set as 100 seconds.

\subsection{Discussion}
\begin{figure} [t!]
    \centering
	{\includegraphics[width=0.9\linewidth]{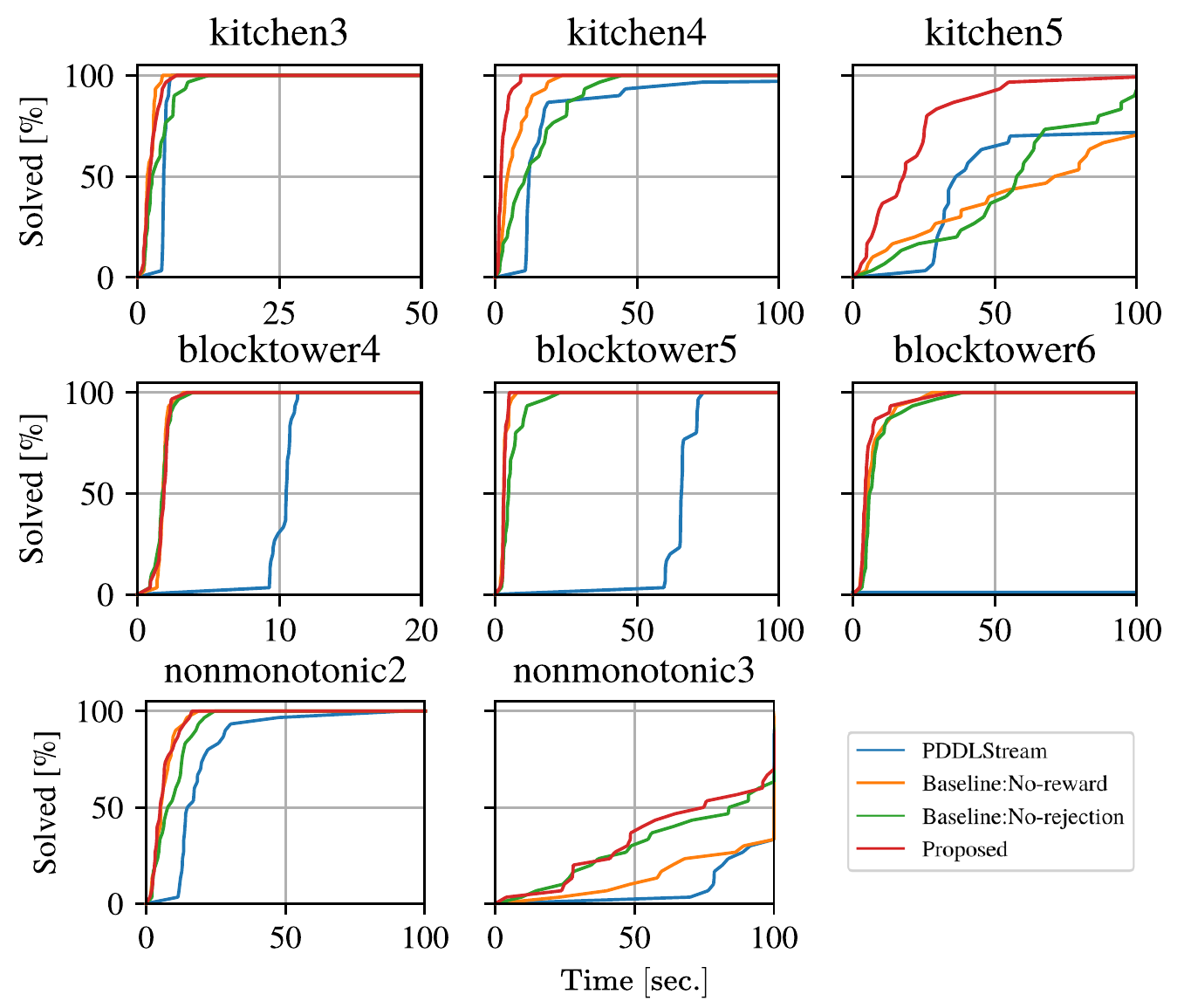}}
	    \vspace{-4mm}
    	\caption{The comparison result of the proposed planner, two baseline planners (no-reward, no-rejection), and PDDLStream. The results are expressed as a cumulative distribution function.}
	\label{fig:validation_result}
	\vspace{3mm}
\end{figure}

Simulation results are presented in Fig. \ref{fig:validation_result} using a cumulative distribution function. In the kitchen domain, it is a better strategy to choose a simple task plan. For example, it is more advantageous to complete the \text{wash}-\text{cook} task in series, because \texttt{sink} has a tight goal region. Our planner showed a success rate of 97\%, while the no-rejection baseline showed a success rate of 90\%, and the PDDLStream and no-reward baseline showed a success rate of around 70\%. 
  
Comparing the performance of the two baselines, the no-reward baseline, which exploits the goal candidate sampling, performed better than the no-rejection baseline in the kitchen 3 and 4 domains. This is intuitively acceptable, as the sparsity of blocks in these domains makes it likely that a randomly sampled abstract plan will be feasible. In contrast, in kitchen 5, the no-rejection baseline, which is the informed planner by a previous reward, performed better. The proposed planner showed the highest performance for all scenarios by utilizing both features.

The effectiveness of rewards can also be seen in non-monotonic domains, in which the blocker blocks must be removed first. 
In the proposed method, the task planner learns through rewards that the action sequence for removing the blocker blocks is more promising. 
As a result, in non-monotonic domains, our proposed method showed the best performance with 66\%. The rejection of using goal candidates did not lead to a significant improvement in performance in this domain. 

Finally, in the blocktower domain, the proposed method including the baselines outperformed PDDLStream. This is because the proposed method solves a much simpler symbolic planning problem compared to the PDDLStream. In fact, \cite{khodeir2021learning} has already reported that the blocktower domain with more than 4 objects is challenging for PDDLStream. In the case of blocktower 6, PDDLStream could not find any solutions, while the proposed method showed a 90\% success rate within 13 seconds. Compared to PDDLStream, our planner stores geometrical planning instances, such as grasp, placement, configuration, and trajectories in a reachability tree instead of including them in the task planning problem. 
In other words, our approach keeps the task planning problem manageable and prevents it from becoming more complicated by sampled instances over time.

\section{Conclusion}
This paper presents a tree-based TAMP solver, inspired by reachability tree-based MMMP solvers. The proposed method samples the abstract action sequence in such a way that the tree expansion does not violate the task constraints defined in the abstract domain. The evaluation of the sampled sequence by the geometric planners allows us to bias the search towards more promising abstract state regions. In addition, a sampling rejection scheme that pre-generates a goal candidate can be used to resolve the tight goal constraints effectively. The comparative study using two baselines and PDDLStream showed the effectiveness of our approach.

\bibliographystyle{IEEEtran}
\bibliography{IEEEabrv, ref}

\end{document}